\documentclass[letterpaper, 10 pt, conference]{ieeeconf}
\IEEEoverridecommandlockouts

\title{\LARGE \bf
Distillation for Efficient Multitask Manipulation Policies \\ via Conditional Flow Matching
}

\author{Shreya Deshmukh, Imen Mahdi, Nick Heppert, Abhinav Valada
\thanks{Department of Computer Science, University of Freiburg, Germany.}%
\thanks{This work was partially funded by the Carl Zeiss Foundation with the ReScaLe project. Nick Heppert is supported by the Konrad Zuse School of Excellence in Learning and Intelligent Systems (ELIZA) through the DAAD programme Konrad Zuse Schools of Excellence in Artificial Intelligence, sponsored by the Federal Ministry of Education and Research.}%
}
\usepackage{amsmath}
\usepackage{amssymb}
\usepackage{graphicx}
\usepackage{xcolor}
\usepackage{tikz}
\usepackage{subcaption}
\usepackage{booktabs}
\usepackage{multirow}
\usetikzlibrary{arrows.meta}
\usetikzlibrary{shadows.blur}
\usetikzlibrary{positioning}
\usepackage[hang,flushmargin]{footmisc}
\usepackage{pgfplots}
\pgfplotsset{compat=1.18}
\usetikzlibrary{decorations.pathreplacing}
\usepackage{float}
\usepackage{stfloats}
\usepackage{placeins}
\usepackage{microtype}
\usepackage{balance}
\usepackage{cite}

\makeatletter
\let\NAT@parse\undefined
\makeatother

\usepackage{hyperref}
\hypersetup{
    colorlinks=true,
    citecolor=black,
    linkcolor=black,
    urlcolor=black
}

\usepackage{cleveref}

\begin{document}

\maketitle
\thispagestyle{empty}
\pagestyle{empty}

\begin{abstract}

Advances in generative modeling have recently been extensively employed in robotics for policy learning. In particular, Conditional Flow Matching (CFM) trained with expert demonstrations has been shown to outperform existing methods on robot manipulation benchmarks. While prior work has mainly focused on single-task settings, we study the problem from a multi-task perspective, as training independent models for each task is computationally expensive. Multi-Task policy learning comes with its own set of challenges, as naively training on a concatenated dataset of demonstrations would either require increased model capacity to accommodate the added complexity or result in drops in performance. We propose to distill knowledge from single-task CFM experts into a shared multi-task policy by transferring their learned velocity fields. We combine this distillation signal with the original CFM objective to retain fidelity to the demonstrations. Experiments on RLBench show that our approach improves multi-task policy performance over naive training while maintaining a fixed model size.

\end{abstract}
\section{Introduction}
Policy learning in robotics has traditionally focused on single-task settings, where a separate policy is trained for each task~\cite{heppert2026scaling}. Recent generative policy learning methods~\cite{zhao2023learning,chi2025diffusion,chandra2025diwa,yang2025covar} have outperformed existing baselines, but storing and training a separate policy for every task becomes increasingly impractical as the number of tasks grows. This motivates learning a single multi-task policy that can reuse knowledge across tasks without increasing model size.

Recent vision-language-action models (VLAs)~\cite{black2024pi_0,bjorck2025gr00t,shukor2025smolvla,argus2025cvla} have demonstrated that large-scale training can enable broad multi-task generalization. However, their substantial model sizes make real-time deployment and task-specific finetuning computationally expensive. In this work, we instead consider multi-task policy learning under a fixed model capacity and limited demonstrations. A key challenge in this setting is negative transfer, where jointly training multiple tasks can degrade performance compared to training separate task-specific policies~\cite{lakkapragada2022mitigating}.

Generative policy frameworks, such as Conditional Flow Matching (CFM)~\cite{chisari2024learning} provide a straightforward framework for learning generative policies. CFM learns a velocity field that transports samples from a simple initial distribution to the target action distribution. While this formulation is effective for single-task policy learning~\cite{chisari2024learning}, naively training one CFM model on multiple tasks introduces conflicting gradients. We propose to address this problem by distilling knowledge from single-task CFM experts into a single multi-task model. A simple way is to utilize the experts' generated actions as the training targets. We go a step further into the generative process by proposing to distill the denoising gradients directly. By using these gradients as soft supervision, the multi-task policy learns shared structure from the experts while remaining anchored to the original demonstrations through the standard CFM objective.

In this work, we study multi-task policy distillation for CFM-based robotic policies. We introduce two distillation variants and analyze their behavior using a toy example and robotic manipulation experiments. Our contributions are:
\begin{enumerate}
\item A framework for distilling multiple single-task CFM experts into a single multi-task policy without increasing student model size.
\item Two distillation strategies based on expert velocity fields and expert-generated target samples.
\item An empirical study and comparison of our method to baselines.
\end{enumerate}
\section{Prerequisites: Conditional Flow Matching}
We adopt Conditional Flow Matching (CFM)~\cite{lipman2022flow} as the general training paradigm for our policies. CFM offers desirable properties, namely its generative nature, ability to model multi-modal distributions, and its simple formulation. CFM frames generation as a continuous transformation, or \emph{flow}: samples are drawn from a simple, known distribution $p_0$ (e.g., normal distribution) and gradually deformed into samples from the data distribution $p_1$ by following a time-dependent vector field $v: \mathbb{R}^d \times [0,1] \rightarrow \mathbb{R}^d$. This vector field induces a flow $\phi: [0,1] \times \mathbb{R}^d \rightarrow \mathbb{R}^d$, defined as the solution to the ordinary differential equation (ODE)
\begin{equation}
    \frac{d}{dt}\phi_t(z) = v\big(\phi_t(z),t\big), \qquad \phi_0(z) = z.
    \label{eq:ode}
\end{equation}
The flow $\phi_t$ thus defines a time-dependent probability density path $p_t : [0,1] \times \mathbb{R}^d \rightarrow \mathbb{R}_{>0}$ interpolating between $p_0$ and $p_1$. Since $p_t$ is not directly available, CFM instead defines it implicitly through a \emph{conditional probability path} $p_t(z \mid z_1)$: a distribution over intermediate samples $z$ constructed for each individual target $z_1 \sim p_1$. In practice, a simple linear conditional path is used, which draws $z_0 \sim p_0$ and $z_1 \sim p_1$ independently and defines the interpolated sample as
\begin{equation}
    z_t = (1-t)\,z_0 + t\,z_1, \qquad t \in [0,1],
    \label{eq:interp}
\end{equation}
with the corresponding ground-truth conditional velocity given in closed form by
\begin{equation}
    v_{\mathrm{gt}}(z_t,t \mid z_1) = \frac{d}{dt}z_t = z_1-z_0.
    \label{eq:vgt}
\end{equation}

We then regress these conditional velocities into a learned velocity field parametrized by a neural network $v_\theta(z_t,t)$, which at inference time is integrated via Eq.~\eqref{eq:ode} to produce samples from $p_1$. During training, we directly construct intermediate samples $z_t$ via Eq.~\eqref{eq:interp} and train the network to match the corresponding ground-truth velocity $v_{\mathrm{gt}}$ from Eq.~\eqref{eq:vgt}, via the following objective:
\begin{equation}
    \mathcal{L}_{\mathrm{CFM}}(\theta)
    =
    \mathbb{E}_{z_0 \sim p_0,\,z_1 \sim p_1}
    \left[
        \left\|
        v_\theta(z_t,t)-(z_1-z_0)
        \right\|^2
    \right],
    \label{eq:cfm_loss}
\end{equation}
where $t$ is sampled from $\mathcal{U}[0,1]$, a uniform distribution.

\section{Multi-Task Policy Distillation with Conditional Flow Matching}
\label{sec:method}
\begin{figure*}[t]
    \centering
    \includegraphics[width=\linewidth, height=0.21\linewidth]{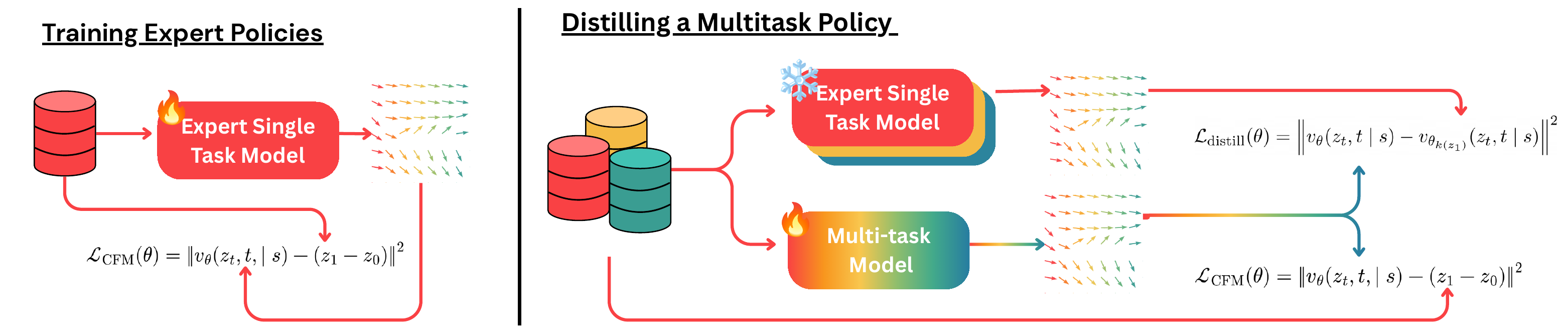}
    \caption{\textbf{Method Overview.} $K$ single-task CFM experts $v_{\theta_k}$ trained on their own datasets $D_k$, and distilled into a single multi-task model $v_\theta$. The aggregated ground truth dataset is used to anchor the final samples into the real data points.}
    \label{fig:method_overview}
\end{figure*}

\subsection{Problem Formulation}
We study a multi-task imitation learning setting, where the goal is to learn a single policy $\pi: \mathcal{S} \rightarrow \mathcal{A}$ capable of solving a set of $K$ tasks $\{T_k\}_{k=1}^{K}$, with $\mathcal{S}$ and $\mathcal{A}$ denoting the state and action spaces, respectively. We assume access to a dataset of $N = \sum_{k=1}^{K} N_k$ expert demonstrations, collected across all tasks, where $D_k = \left\{ \left(s^{(i)},a^{(i)}\right) \right\}_{i=1}^{N_k}$ denotes the $N_k$ demonstrations available for task $T_k$. We denote the union of all task datasets by $D = \bigcup_{k=1}^{K}D_k$.

In this work, we cast policy learning as an instance of the CFM framework introduced above, where the flow variable $z$ corresponds to an action trajectory (or action chunk) $a$. Each sampled expert demonstration is a pair $(s,a)\sim D$; we set the flow target $z_1 = a$ and condition the learned velocity field on the paired state, $v_\theta(z_t,t \mid s)$. For notational clarity, in what follows we omit this conditioning on $s$ from the sampling distributions themselves and simply write $z_1\sim p_1$, with the state dependence remaining implicit in every $v_\theta(z_t,t \mid s)$ term.

\subsection{Multi-Task policy distillation}
Training single-task expert policies with conditional flow matching has been shown to achieve strong performance on robotic manipulation benchmarks. However, as the number of tasks $K$ grows, training and storing a separate expert per task becomes increasingly inefficient. This motivates learning a single \emph{multi-task} policy that remains compact while matching the performance of the per-task experts. Achieving this requires explicit cross-task knowledge transfer, since robotic manipulation tasks commonly share an overlapping operational space and similar trajectory behavior (e.g., reaching, grasping, smooth motions), while differing mainly in fine-grained, task-specific behavior (e.g., which objects to manipulate). Naively training a single CFM policy on the union of all task datasets does not enforce this shared structure: it either requires increasing model capacity to absorb the combined complexity of all $K$ target distributions, or leads to degraded per-task performance, since the conditional velocity targets $v_{\mathrm{gt}}=z_1-z_0$ used in Eq.~\eqref{eq:cfm_loss} can exhibit high variance across tasks and demonstrations.

Our intuition is that a CFM velocity field trained on a single-task already learns probability paths that map to a close representation of the target distribution of that task. The learned velocity fields $v_{\theta_k}$ could then be used as a stronger guidance signal than the raw straight-line target $v_{\mathrm{gt}}$. We therefore first train $K$ single-task expert policies, each following the standard CFM training objective on its own dataset $D_k$.

To train a multi-task $v_\theta(z_t,t \mid s)$, we then distill the corresponding expert's prediction at the same interpolated point $z_t$. For each sampled demonstration $z_1\sim p_1$, the task associated with the demonstration determines which expert provides the teacher signal:
\begin{equation}\scriptsize
    \mathcal{L}_{\mathrm{distill}}(\theta)
    =
    \mathbb{E}_{z_0 \sim p_0,\,z_1\sim p_1}
    \left[
        \left\|
        v_\theta(z_t,t \mid s)
        -
        v_{\theta_{k(z_1)}}(z_t,t \mid s)
        \right\|^2
    \right],
    \label{eq:distill_loss}
\end{equation}
where $k(z_1)$ denotes the task index associated with the sampled demonstration $z_1\in D_k$. Thus, the same demonstration used to construct the CFM target in Eq.~\eqref{eq:cfm_loss} determines the corresponding expert teacher. Since $v_{\theta_k}$ was trained on task $T_k$ alone, it provides a smoother and less noisy target than $v_{\mathrm{gt}}$, particularly useful during the early stages of the integration. 
However, relying purely on $\mathcal{L}_{\mathrm{distill}}$ can be detrimental in the later integration steps ($t$ close to $1$), where fine, task-specific behavior needs to be captured precisely, and any residual error in the expert prediction would otherwise be propagated into the student. We therefore anchor the multi-task model to the original data distribution by combining Eq.~\eqref{eq:cfm_loss} (evaluated for the multi-task $v_\theta$) with the distillation loss:
\begin{equation}
    \mathcal{L}_{\mathrm{total}}(\theta)
    =
    (1-\lambda)\,\mathcal{L}_{\mathrm{CFM}}(\theta)
    +
    \lambda\,\mathcal{L}_{\mathrm{distill}}(\theta),
        \label{eq:total_loss}
\end{equation}
where $\mathcal{L}_{\mathrm{CFM}}(\theta)$ is Eq.~\eqref{eq:cfm_loss} applied to the multi-task model $v_\theta(z_t,t \mid s)$ across all tasks, and $\lambda \in [0,1]$ is a fixed scalar weighting the contribution of the distillation term relative to the ground-truth CFM loss.


\subsection{Distilling Learned Target Distribution vs.\ Velocity Vector Fields}
An alternative to distilling the expert velocity field is to instead use the expert policies to augment the data points from the \emph{target distribution} $p_1$ itself. Since each expert $v_{\theta_k}$ is a generative model that has learned to approximate the target distribution of task $T_k$ well, it can be used to synthesize additional plausible expert-like actions beyond the fixed set of $N_k$ demonstrations, thereby increasing the diversity and effective coverage of the training targets at negligible additional data-collection cost.

Concretely, for a given state $s$ observed in $D_k$, we sample a noise vector $z_0 \sim p_0$ and numerically integrate the frozen expert ODE forward in time with $v=v_{\theta_k}$, i.e.\ we evaluate the induced expert flow at $t=1$,
\begin{equation}
    \hat{a}
    =
    \phi_1^k\left(z_0 \mid s\right),
    \label{eq:expert_rollout}
\end{equation}
where $\phi_t^k(\cdot \mid s)$ denotes the solution of Eq.~\eqref{eq:ode} with velocity field $v_{\theta_k}(\cdot,\cdot \mid s)$ and initial condition $\phi_0^k(z_0 \mid s)=z_0$. Repeating this procedure for multiple noise samples $z_0$ per state $s$ yields a set of synthetic action targets $\hat{a}$, which we add to $D_k$ as additional $(s,\hat{a})$ pairs, augmenting the original expert demonstrations.

However, this approach also inherits errors from the learned expert. In contrast to the original demonstrations, a generated target $\hat{a}$ is not ground truth but a sample from the learned expert distribution. Thus, any modeling error in the learned expert is directly transferred to the velocity targets used to train the student:
\begin{equation}
    v_{gt} = z_1-z_0 = \phi_1^k\left(z_0 \mid s\right) - z_0.
\end{equation}

Consequently, although expert-based data augmentation can increase the coverage of the target distribution, its effectiveness depends strongly on the quality of the learned single-task experts.
\section{Experimental Evaluation}
In this section, we aim to answer the following research questions:
\begin{enumerate}
    \item What is the intuition behind why distilling knowledge from single-task expert policies helps multi-task learning, and what are the limitations and benefits of each proposed variant?
    \item Does distilling from single-task experts enable positive cross-task knowledge transfer on a real manipulation benchmark compared to naive multi-task training?
    \item Between distilling the experts' velocity fields and distilling their generated target samples, which better balances transfer and fidelity to the true action distribution?
    \item How does the benefit of distillation evolve as the number of tasks $K$ grows, and how sensitive is performance to the distillation weight $\lambda$?
\end{enumerate}

\subsection{Toy Example: 1D Gaussian Distributions}
\begin{figure*}[t]
\centering
\includegraphics[width=\linewidth]{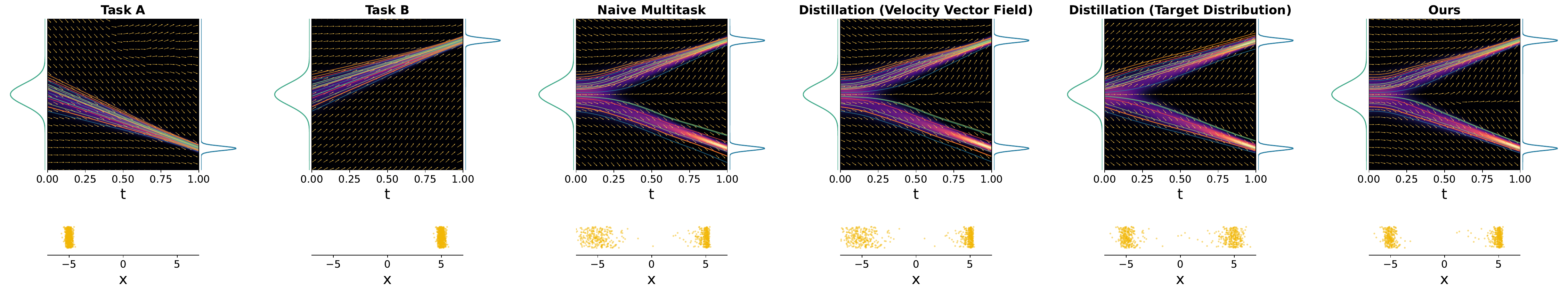}
\caption{\textbf{Toy 1D example}. We visualize the learned flow that transports an initial distribution $p_0 \sim \mathcal{N}(0,1)$ to two well-separated target modes: Task A, $\mathcal{N}(\mu{=}{-}5, \sigma{=}0.2)$, and Task B, $\mathcal{N}(\mu{=}5, \sigma{=}0.2)$. Each panel shows the learned probability path $p_t(x)$ as a heatmap over $(t,x)$, the corresponding velocity field as arrows, and several integrated sample trajectories from $t{=}0$ to $t{=}1$. The source density $p_0$ is shown on the left and the target density $p_1$ on the right, with a strip of the flow's actual final samples shown underneath. \emph{Task A} and \emph{Task B} are single-task experts, each trained with the standard CFM objective on its respective target mode, and learn a single clean, low-noise transport path. \emph{Naive Multi-Task} is trained on the pooled Task A and Task B data using the same objective. Its early-time velocity field is shared across both tasks, but its final samples exhibit noticeably higher variance, as jointly matching two disjoint targets provides a harder and noisier learning signal. \emph{Distilling Velocity Vector Fields} regresses the multi-task model only against the pooled experts' velocity predictions. It exhibits similar early-time sharing, but because it is never anchored to the real target samples, its final distribution does not fully recover either mode. \emph{Distilling the Experts' Learned Target Distributions} instead trains on samples generated by the experts, producing final distributions closer to the true targets, but its velocity field inherits noise and bias from the experts' predictions. \emph{Ours} combines both signals: it distills the experts' velocity fields while anchoring the final flow to the real data, yielding the closest match to each target mode.}
\label{fig: toy_example}
\end{figure*}
To build intuition for how velocity-field distillation can help before moving to the full manipulation benchmark, we construct a minimal setting in which each task corresponds to a unimodal Gaussian. Two single-task experts are trained with the standard CFM objective to transport an initial distribution $p_0=\mathcal{N}(0,1)$ to two parameterized target distributions $\mathcal{N}(\mu,\sigma)$. Fig.~\ref{fig: toy_example} compares different methods trained to model the aggregated samples from both tasks.

We first train a naive multi-task model on the pooled data using the standard CFM loss. While the learned velocity field is shared across both tasks early in the flow, the final induced distribution exhibits higher variance, since jointly optimizing for both tasks produces a noisier learning signal. Distilling the velocity fields without directly regressing against the real target samples exhibits similar early-stage sharing. However, because the student is never explicitly optimized to match the target distribution, its final samples do not fully recover either mode. Training instead on samples generated by the experts produces final distributions that are closer to the true targets, but the resulting velocity field is more sensitive to errors and variance inherited from the experts. Our full objective, described in Eq.~\eqref{eq:total_loss}, combines both signals: distilling the experts' velocity fields while anchoring the flow to the real data, thereby producing samples that most closely match the target distributions.

\subsection{Benchmark}
We evaluate our method on RLBench~\cite{james2020rlbench}, a popular robot learning benchmark built around a 7-DoF Franka Emika Panda arm. Following prior work on single-task policy learning~\cite{chisari2024learning}, we consider a set of $K=8$ manipulation tasks: \textit{unplug charger}, \textit{close door}, \textit{open box}, \textit{open fridge}, \textit{take frame off hanger}, \textit{open oven}, \textit{put books on bookshelf}, and \textit{take shoes out of box}. For each task, we collect 100 expert demonstration trajectories using RLBench's automatic expert demonstration execution, consisting of full robot states and camera observations. At evaluation time, we roll out each policy for 100 episodes per task, with randomized task setup. We report the success rate (SR) per task, as well as the mean SR across all $K$ tasks.

\subsection{Baselines}
\begin{enumerate}
    \item \textbf{Expert Policies}: For each task $T_k$, we train a single-task policy on dataset $D_k$ using the loss described in Eq.~\eqref{eq:cfm_loss}. This is the same policy we distill from in our method.
    \item \textbf{Naive Multi-Task Policy}: This policy shares the same architecture as the single-task policies, but is trained on the combined dataset of all tasks, $D=\bigcup_{k=1}^{K} D_k$, using the same loss. We do not provide task conditioning as input, since we assume the tasks are diverse enough for the task identity to be inferred directly from the state $s$.
    \item \textbf{Knowledge Amalgamation}: Proposed by \cite{luo2019knowledge}, this technique aligns feature spaces across tasks by introducing a probing layer that matches learned task-specific features, thereby enabling knowledge transfer across tasks.
\end{enumerate}

We compare these baselines against two variants of our method, described in Sec.~\ref{sec:method}: learned target-distribution distillation and velocity-field distillation with the loss described in Eq.~\eqref{eq:total_loss}. We adopt the PointFlowMatch architecture~\cite{chisari2024learning} as the base model for both the single-task experts $v_{\theta_k}$ and the multi-task model $v_\theta$. Point cloud observations are encoded with a modified PointNet backbone (T-nets removed), and the resulting visual features are concatenated with the robot's proprioceptive state to form the conditioning vector, injected at the bottleneck of a conditional 1D U-Net that predicts the velocity field. Following PointFlowMatch, inference is performed by numerically integrating the learned ODE with a fixed-step solver over a discretized number of integration steps. Notably, the multi-task model does not receive an explicit task identifier as input: it is conditioned on the observation $s$ alone, using the same architecture as the single-task experts. We rely on the point cloud and proprioceptive observations being sufficiently distinct across tasks that the task identity is implicitly encoded within $s$, allowing a single shared PointNet backbone and U-Net to disambiguate and handle all $K$ tasks without a task label at either training or inference time. We set $\lambda=0.5$ in all experiments unless stated otherwise. Each model is optimized using AdamW optimizer with a learning rate of $3e^{-5}$ and a weight decay of $1 e^{-6}$.

\subsection{Evaluation Results}
\begin{table*}[t]
\centering
\caption{Per-task success rates (\%) across 8 RLBench tasks with baselines: Single-Task, Naive Multi-Task, Amalgamation and both variants of our method. Best result per task is \textbf{bolded}.}
\label{tab:8task_main_table}
\footnotesize
\setlength{\tabcolsep}{6.5pt}
\begin{tabular}{l c c c c c c c c c}
\toprule
 & \shortstack{unplug \\ charger} 
 & \shortstack{close \\ door} 
 & \shortstack{open \\ box} 
 & \shortstack{open \\ fridge} 
 & \shortstack{take frame \\ off hanger} 
 & \shortstack{open \\ oven} 
 & \shortstack{put books \\ on bookshelf} 
 & \shortstack{take shoes \\ out of box} 
 & \textbf{Avg.} \\
\midrule
Expert Policy~\cite{chisari2024learning}    & 70.0          & 67.0          & 99.0  & 46.0          & 35.0          & 68.0          & 58.0          & 72.0          & 64.4 \\
Naive Multi-Task     & \textbf{85.0} & 45.0          & 97.0           & 31.0          & 36.0          & 74.0          & 69.0          & 54.0          & 61.4 \\
Amalgamation~\cite{luo2019knowledge}   & 74.0          & 75.0          & \textbf{100.0} & 50.0          & 41.0          & \textbf{85.0} & 74.0          & \textbf{89.0} & 73.5 \\
Ours (Distilling Learned Target Distribution)   & 78.0          & 42.0 & 95.0           & 53.0 & 66.0 & 81.0          & \textbf{75.0} & 71.0          & 70.0 \\
Ours (Eq.~\eqref{eq:total_loss})   & 79.0          & \textbf{84.0} & 83.0           & \textbf{77.0} & \textbf{77.0} & 79.0          & 74.0 & 83.0          & \textbf{79.5} \\
\bottomrule
\end{tabular}
\end{table*}
We summarize the results in Table~\ref{tab:8task_main_table}. It is important to note that the expert policies converge to substantially different levels of performance, with success rates ranging from 35\% to 99\%. Additionally, some tasks share some common behavior such as opening tasks but differ in the target object. This variation is important for our study, as we aim to investigate which architectures and learning methods can enable poorly performing tasks to benefit from knowledge acquired from other tasks. The naive multi-task policy, trained on the aggregated dataset, underperforms the average performance of the single-task experts. This degradation is expected, as the increased task complexity reduces the effective capacity available to each individual task.

Knowledge Amalgamation enforces sharing in the learned feature space through an additional probing layer, resulting in improved multi-task performance. Notably, its performance exceeds that of the single-task average, suggesting that it encourages the learning of general behaviors that benefit multiple tasks. Distilling the expert-generated trajectories further improves performance over the naive multi-task baseline; however, the noise introduced by the expert policies still limits the final performance.

Our full objective, defined in Eq.~\eqref{eq:total_loss}, achieves the best overall performance, reaching an average success rate of 79.5\%. It leverages both the learned velocity fields from the expert policies and an anchoring term that encourages the flow toward the real data points. Most notably, both variants of our method improve over the average expert performance, similarly to Knowledge Amalgamation. This result demonstrates that the soft labels provided by expert policies can facilitate effective knowledge transfer across tasks. This is most apparent in the least performing tasks such as \textit{open fridge} that benefit the most from the other similar tasks such as \textit{open box} that performs well, thus increasing SR from 46\% to 77\%.

\subsection{Ablations}
\begin{figure}[t]
    \centering
    \includegraphics[width=0.95\linewidth]{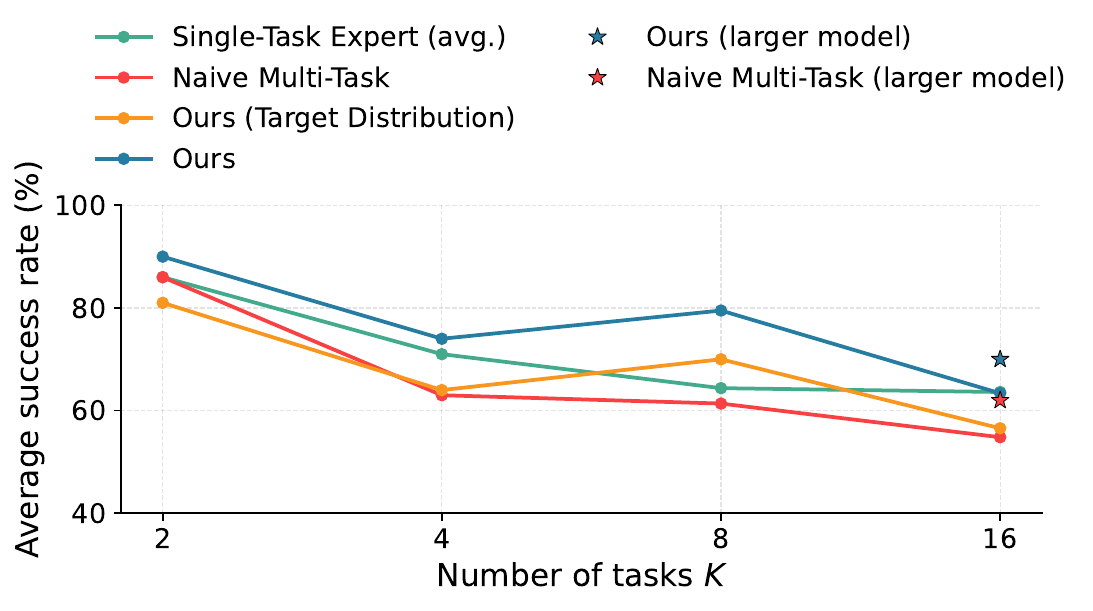}
    \caption{\textbf{Scalability ablation.} Average success rate as a function of the number of tasks $K \in \{2,4,8,16\}$ for the naive multi-task baseline, our distilled student (both variants), and the corresponding average of $K$ single-task experts. A larger student variant is additionally reported at $K=16$ (star markers) to disentangle capacity limitations from our approach itself.}
    \label{fig:scalability}
\end{figure}

\paragraph{Scalability}
To assess how our method scales with the number of tasks, we train the naive multi-task baseline and both variants of our model on progressively larger task sets, ranging from $K=2$ to $K=16$. We compare their average success rate (SR) against the corresponding average performance of the $K$ single-task experts; the results are shown in Fig.~\ref{fig:scalability}.

The naive multi-task baseline consistently underperforms the average of the single-task experts, failing to achieve positive knowledge transfer across tasks even as the model capacity is increased (at $K=16$). Simply distilling the experts' final generated trajectories also does not reliably improve performance, as this approach is sensitive to errors in the experts' generated trajectories. In contrast, training our model with the combined objective of matching the expert velocity fields while anchoring to the real data consistently improves performance, both over the naive multi-task baseline and, for $K \leq 8$, over the average single-task expert. This suggests that our distillation objective enables the model to exploit shared structure across tasks and make more effective use of its capacity.

At $K=8$, the advantage over the single-task expert average becomes particularly pronounced, with our method achieving 79.5\% SR compared to 64.4\%. The increased task diversity provides more opportunities for positive transfer, while the lower average performance of the individual experts leaves greater room for improvement through cross-task knowledge sharing. At $K=16$, however, performance falls back to approximately the single-task expert average, suggesting that the student has reached its capacity limit. To distinguish this capacity limitation from a fundamental limitation of our distillation approach, we repeat the $K=16$ experiment with a larger student architecture. Performance recovers and follows the same positive trend observed for $K \leq 8$. This confirms that our distillation objective can effectively exploit additional capacity to encode shared cross-task behavior, rather than the observed gains being attributable solely to easier task compositions or dataset effects at smaller $K$.

\paragraph{Sensitivity}
We study the sensitivity of our method to the fixed weighting term $\lambda$ in Eq.~\eqref{eq:total_loss}, which balances the distillation loss $\mathcal{L}_{\mathrm{distill}}$ against the ground-truth CFM loss $\mathcal{L}_{\mathrm{CFM}}$. We sweep $\lambda \in \{0, 0.25, 0.5, 0.75, 1.0\}$ and report the average success rate across all $K=8$ tasks in Table~\ref{tab:lambda_sensitivity}. Note that $\lambda=0$ reduces Eq.~\eqref{eq:total_loss} to $\mathcal{L}_{\mathrm{CFM}}$ alone applied to the pooled dataset, exactly recovering the naive multi-task baseline. We find that $\lambda = 0.5$, which weighs both loss terms equally, achieves the best performance (79.5\%), while values skewed toward either extreme underperform: with no distillation signal at all ($\lambda=0$) performance is worst (61.4\%), a low but nonzero $\lambda$ still under-utilizes the smoother distillation signal from the expert velocity fields ($\lambda=0.25$: 69.0\%), while a high $\lambda$ over-relies on the expert prediction at the expense of matching the ground-truth target actions ($\lambda=0.75$: 68.0\%, $\lambda=1.0$: 65.4\%). Yet, distilling the experts' velocity fields leads to improved performance compared to the single-task average.

\begin{table}[t]
\centering
\caption{\textbf{Sensitivity to $\lambda$.} Average success rate (\%) across 8 RLBench tasks for varying values of the distillation weight $\lambda$ in Eq.~\eqref{eq:total_loss}, alongside the single-task expert baseline for reference. $\lambda=0$ reduces Eq.~\eqref{eq:total_loss} to the naive multi-task objective.}
\label{tab:lambda_sensitivity}
\footnotesize
\setlength{\tabcolsep}{2pt}
\begin{tabular}{l c c c c c c}
\toprule
 &  & \multicolumn{5}{c}{$\lambda$} \\
\cmidrule(lr){3-7}
 & Single-Task & 0 & 0.25 & 0.5 & 0.75 & 1.0 \\
\midrule
Avg. SR (\%) & 64.4 & 61.4 & 69.0 & \textbf{79.5} & 68.0 & 65.4 \\
\bottomrule
\end{tabular}
\vspace{-0.3cm}
\end{table}
\section{Limitations and Future Work}
In this work, we have used a fixed distillation weight $\lambda$. In the following
work, we aim to investigate a time-dependent schedule $\lambda(t)$ favoring distillation early in the flow and ground-truth supervision later. Finally, our evaluation is limited to simulated RLBench tasks with single-arm manipulation. Next steps would include validating our method on more benchmarks and on real robot setups.

\section{Conclusion}
We presented a method for distilling single-task expert policies, trained with conditional flow matching, into a single compact multi-task policy for robotic manipulation. Instead of naively training on the union of all task datasets, we use the well-structured velocity fields learned by single-task experts as a training signal for a shared student, either by distilling the expert velocity field directly or by using the experts to augment the target action distribution. Our experiments on RLBench show that both variants outperform a naive multi-task baseline, and even the average of the single-task experts themselves, with the benefit growing as more tasks are added until the student saturates its model capacity. These results suggest that our distillation objective enables effective cross-task knowledge transfer within a single shared policy.

\balance
\bibliographystyle{IEEEtran}
\bibliography{bibliography}

\end{document}